# Entropy and Belief Networks

Norman C. Dalkey
Department of Computer Science
University of California, Los Angeles
Los Angeles, CA 90024

## Abstract

The product expansion of conditional probabilities for belief nets is not maximum entropy. This appears to deny a desirable kind of assurance for the model. However, a kind of guarantee that is almost as strong as maximum entropy can be derived. Surprisingly, a variant model also exhibits the guarantee, and for many cases obtains a higher performance score than the product expansion.

## 1 Introduction

In a previous report I showed that the product expansion of conditional probabilities for belief nets is not maximum entropy [Dalkey, 88]. The product expansion is the established procedure for extending probabilistic information on subsets of variables to complete distribution for belief networks (causal networks, Bayesian networks, influence diagrams) [Pearl, 88; Shacter, 86; Lauritzen and Spiegelhalter, 88]. If the product expansion were maximum entropy (maxent for short), it would carry a strong guarantee, namely, that even if the assumptions of conditional independence that underly the product form were incorrect, the expected (logarithmic) score of the computed probability distribution would be just as good as if the assumptions were correct [Dalkey, 85]. Since the inputs to many applications of belief networks rely heavily on fallible expert judgment, the guarantee would be a strong additional assurance for accepting the models.

Although the product expansion is not maxent, it will be shown below that it carries a kind of guarantee that is almost as strong. The conditional logarithmic score of the product expansion, given an hypothetical correct distribution, is independent of the correct distribution; thus, even if the correct distribution does not exhibit the conditional independencies assumed in the product form, the conditional score is guaranteed. As a bonus, the conditional score can be computed locally, without knowing the correct distribution.

As a bit of serendipity, in comparing the performance of the product expansion with the performance of a variant model it turned out that the variant model also has the guarantee property, and in addition, for a wide class of cases literally outperforms the product expansion (to be called hereafter the *standard model.*)

The variant model (described more fully below) appears to have a different underlying framework than the conceptual standard model, and its appropriateness for applications remains to be investigated.

## 2 Webs

It is convenient to couch the analysis in terms of a structure I call a *web* that is slightly more general than the usual belief network. Consider a set of variables $X = (X_1, ..., X_n)$ and a set of subsets of these variables $C = (C_1, ..., C_m)$. $C$ is called a *structure* and the subsets $C_i$ are called *components* of the structure. A component $C_i$ is called *terminal* if it contains variables that do not occur in any other component of $C$. The variables unique to $C_i$ are called the tail $T_i$ of $C_i$ and the non-unique variables are called the overlap $O_i$. A web is defined by the recursion:

1. $C$ contains a terminal component $C_i$.
2. If $C$ contains more than one component, then $C - \{C_i\}$ is a web.

From the definition, a web permits an *unpacking*, i.e., a successive removal of components terminal with respect to the remaining structure, which exhausts all components. An unpacking arrays the components in a linear order where at each stage $i$, the current terminal $C_i = (T_i \mid O_i)$ has a tail $T_i$ of variables not contained in the remaining components and an overlap $O_i$ of variables contained in some remaining components. A DAG (directed acyclic graph), the underlying structure for most belief networks, corresponds to a web for which, in a given unpacking, the tails $T_i$ are unit sets. The overlap $O_i$ corresponds in this case to the parents of the variable $T_i$. A hypertree (or tree for short)corresponds to a web such that there is an unpacking in which the overlaps $O_i$ are contained in a single component $C_j$ in the remaining structure.

A probability system $PC = \{P(C_1), ..., P(C_m)\}$ is a set of joint probability distributions $P(C_i)$ on the variables in each component $C_i$. For a given unpacking, the conditional distributions

$$P(T_i \mid O_i) = \frac{P(C_i)}{\sum_{-O_i} P(C_i)} = P(C_i)/P(O_i)$$

correspond to the conditional probabilities of the variables given parents in a belief network. The notation $-O_i$ designates summation over all the variables not in $O_i$.

A system $PC$ is called *consistent* if there is a probability distribution $P(X)$ on the variables $X$ such that

$$\sum_{C_i} P(X) = P(C_i) \quad (1)$$

Similarly, a system $PC$ is called *conditionally consistent* if there is a distribution $P(X)$ such that

$$\frac{\sum_{-C_i} P(X)}{\sum_{-O_i} P(X)} = P(T_i \mid O_i) \quad (2)$$

The product extension $P^x(X)$ of a system $PC$ is defined as

$$P^x(X) = \frac{\Pi_C P(C_i)}{\Pi_O P(O_i)} = \Pi_C P(T_i \mid O_i) \quad (3)$$

It is straightforward to show that $P^x(X)$ is a probability distribution on $X$ and is conditionally compatible with $PC$; i.e., it fulfills (2) for all $C_i$. $P^x$ is the standard model for belief networks generalized to webs.

## 3  Logarithmic Scores

A scoring rule is a reward function $S(P, X_i)$ which assigns a payoff (score) to the distribution $P$ if event $X_i$ occurs. A proper score is one which fulfills the condition

$$\sum_X P(X) S(P, X_i) \geq \sum_X P(X) S(Q, X_i) \quad (4)$$

where $Q$ is any distribution on $X$. For subjective probabilities, (4) states the so-called "honesty condition"; if an estimator believes $P$, his subjective expected score is greater if he reports $P$ than if he reports any other distribution $Q$. For objective probabilities, (4) requires that the objective expectation be a maximum when the score is based on the correct distribution $P$ (the distribution that determines the expectation.)

There is a large family of proper scores [Dalkey, 85]. For the present analysis, the relevant figure of merit is the logarithmic score, $S(P, X_i) = \log P(X_i)$. The expectation of the logarithmic score $\sum_X P(X) \log P(X)$ is the negative of the Shannon entropy [Shannon, 49]. It is closely related to notions of information in information theory and is central to the form of inductive inference known as the maximum entropy method [Jaynes, 68]; [Shore and Johnson, 81].

Let $G(P) = \sum_X P(X) \log P(X)$ and $G(P, Q) = \sum P_X(X) \log Q(X)$

(4) can be written as

$$G(P) \geq G(P, Q)$$

$G(P, Q)$ is the relative score of $Q$ given $P$. It can be interpreted as the expected score if $Q$ is asserted, but $P$ is the case, i.e., $P$ determines the expectation.

If a system $PC$ is consistent, then in general there will be a set of probability distributions $K$ which fulfill (1). Knowing $PC$, the correct probability distribution is in $K$, but otherwise is not known. Consider the distribution $P'$ with maximum entropy in $K$, i.e., $P' = arg\ min_K G(P)$. It can be shown that for any distribution $P$ in $K$,

$$G(P) \geq G(P, P') \geq G(P') \quad (5)$$

that is, the expected score $G(P')$ of the maximum entropy distribution is guaranteed in the strong sense that if any other distribution in $K$ is the actual distribution, the relative score $G(P, P')$ will be at least as great as $G(P')$ [Dalkey, 85].

$P^x$ is not maximum entropy in $K$. However, consider the relative score

$$G(P, P^x) = \sum_X P(X) \log \frac{\Pi_C P(C_i)}{\Pi_O P(O_i)}$$

with $P$ in $K$. The log of the product generates a sum of logs and since $P$ in in $K$, it sums to $P(C_i)$ and $P(O_i)$ for each $i$. Thus,

$$G(P, P^x) = \sum_C G(P(C_i)) - \sum_O G(P(O_i)) \quad (6)$$

The relative expected score of $P^x$ given any $P$ in $K$ is the sum of the expected scores of the marginal on the components minus the sum of the expected scores on the overlaps. In short, $P$ has disappeared in the terms on the right, and $G(P, P^x)$ is independent of $P$. Put in other words, even if the correct probability distribution does not exhibit the conditional independencies appropriate for the product expansion, the product expansion has a guaranteed expection, no matter what the correct distribution.

(6) does not allow asserting the stronger outcome of (5), namely $G(P, P^x) \geq G(P^x)$. In general $G(P^x)$ may be either greater or smaller than $G(P, P^x)$. In words, $P^x$ may "promise" a higher expectation than that guaranteed by the $G(P, P^x)$. However, as (6) shows, the guaranteed expectation can be computed locally, and no ambiguity need arise. In the cases that $G(P, P^x) \geq G(P^x)$, there is the added pleasant irony that $P^x$ is actually better (has a higher assured expectation) than it purports to have.

As a simple example, consider the belief net of Fig. 1.

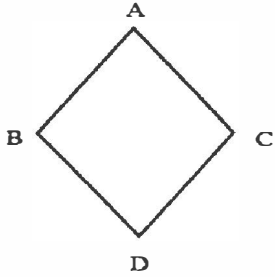

Figure 1

with probabilities $P(A) = .5, P(B) = .6, P(C) = .3, P(AB) = .4, P(AC) = .2, and P(D|BC)$ given by the table

| B | C | P(D\|BC) |
|---|---|---|
| 1 | 1 | .8 |
| 1 | 0 | .6 |
| 0 | 1 | .4 |
| 0 | 0 | .2 |

$$P^x(ABCD) = \frac{P(AB)P(AC)P(BCD)}{P(A)P(BC)} = \frac{P(AB)P(AC)P(D|BC)}{P(A)}$$

According to (6) $G(P, P^x) = G(P(AB)) + G(P(BCD)) - G(P(A)) - G(P(BC)))$. Performing the arithmetic, $G(P, P^x) = -2.48778$. In this case $G(P^x) = -2.45036$, so the guaranteed expectation is a little less than what $P^x$ "promises".

## 4 An Alternative Model

The numerical values of the logarithmic score are not easily grasped intuitively. We can contrast the expected score with that of a completely uninformative $PC$ (uniform distribution on $X$). Since there are 16 joint states for $ABCD$, the uninformative distribution has an expected score of $log\ 1/16 = -2.77259$, definitely less than $G(P, P^x)$; but I'm not sure that helps much in seeing what the guarantee does for you.

It occurred to me that it might be illuminating to contrast the performance of the standard model with a variant model using the same inputs. Clearly, the alternative model shouldn't be completely stupid. An alternative was suggested by the fact that there are two different ways to conceive of the overlaps in a structure $C$. In the standard model, the overlaps in Fig. 1 are $A$ and $BC$. However, considering the overlaps as set intersections, we have the situation in Fig. 2. $A$ is the overlap between $AB$ and $AC$, but $AB$ overlaps $BCD$ only in $B$, and $AC$ overlaps $BCD$ only in $C$. Since the overlaps are what provide the basic interactions between the components, it seemed that a non-dumb model might take the form

$$P*(X) = \frac{\Pi_C P(C_i)}{\Pi_{O*} P(O_i*)} k \quad (7)$$

where $O*$ is the set of overlaps defined as set intersections. The normalizing factor $k$ is included, since the ratio of the two products need not add to 1. It is clear the set $O*$ is more extensive than the set $O$, and in fact, consists of subsets of members of $O$.

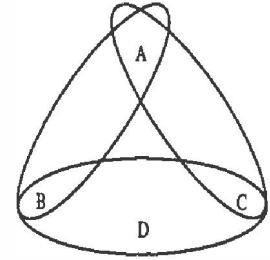

Figure 2

If we determine the relative score of $P*$, with the $P$ in $K$, we obtain

$$G(P, P*) = \sum_C G(P(C_i)) - \sum_{O*} G(P(O_i*)) + log\ k \quad (8)$$

Once again $P$ has disappeared from the right hand side, and we can assert that $P*$ has a guaranteed expectation, whatever the correct probability $P$ might be.

(8) is quite similar to (7). The first term on the right is the same for both. The second terms are analogous, but involve different sets of overlaps. (8) has the additional term $log\ k$. Overlooking $log\ k$ for the moment, the comparison of the second terms is straightforward. In (8) the overlaps are exclusive subsets of the overlaps in (7), and thus we can assert that $\sum_O G(P(O_i)) \geq \sum_{O*} G(P(O_i))$, following the general rule that $G(P(AB)) \geq G(P(A)) + G(P(B))$. Thus, aside from $log\ k$, a larger sum is being subtracted in (7), and $G(P, P^x) \geq G(P, P*)$. In addition, if $k$ is greater than 1, we can assert the inequality a-fortiori.

In the case of our example above, $G(P, P*) = -2.39226$, a definite improvement over $G(P, P^x)$. In this case $k$ is greater than 1. However, when the inputs are modified so that $k$ is less than 1, it remains the case that $G(P, P*)$ beats $G(P, P^x)$.

Thus, we can assert that the alternative model retains the desired guaranteed expectation property, but in addition for a wide range of cases performs better with the logarithmic score.

The alternative model has some other things going for it. It is applicable to structures other than webs,

and retains the guaranteed expectation property in the more general applications. It gives the same result for trees as the standard model since the overlaps for trees are the same as the set intersections.

The good showing of the alternative model was completely unexpected. It is not conditionally consistent, and thus might not be useful as a representation of causal relations. However, it might be relevant for fields such as demographic analysis where the relationships are not overtly causal.

To sum up: Although the standard model is not maximum entropy, it does have a solid guarantee which gives assurances that assumptions of conditional independence are not risky. And there is an alternative model which has the same guarantee and outperforms the standard model, at least for a wide range of cases.

### References

Dalkey, N. C. (1985), "Inductive Inference and the Maximum Entropy Principle," in C. Ray Smith and W. T. Grandy (eds.) *Maximum-Entropy and Bayesian Methods in Inverse Problems*, D. Reidel.

Dalkey, N. C. (1988), "Models vs. Inductive Inference for Dealing with Probabilistic knowledge," in J. F. Lemmer and L. N. Kanal (eds.) *Uncertainty in Artificial Intelligence 2*, Elsevier.

Jaynes, E. T. (1968), "Prior Probabilities," *IEEE Transactions Syst. Sci. Cyber.*, SSC-4, 227-241.

Lauritzen, S. L. and Spiegelhalter, D. J. (1988), "Local Computations with Probabilities on Graphical Structures and their Applications to Expert Systems," *J. Roy. Statist. Soc.*, B, 50, 157-224.

Pearl, J. (1988), *Probabilistic Reasoning in Intelligent Systems*, Morgan Kaufman.

Shacter, R. (1986), "Evaluating Influence Diagrams" *Operations Research*, 34 (no. 6), 871-882.

Shannon, C. E. and Weaver, W. (1949), *The Mathematical Theory of Communication*, Univ. of Illinois Press.

Shore, J. E. and Johnson, R. W. (1981), "Properties of Cross-Entropy Minimization," *IEEE Transactions on Information Theory*, IT-227, 472-482.